\documentclass[%
 preprint,
 amsmath,amssymb,
]{revtex4-1}

\usepackage{graphicx}
\usepackage{amssymb}
\usepackage{setspace}
\usepackage[utf8]{inputenc}
\usepackage{array}
\usepackage{natbib}
\usepackage{amsmath}
\usepackage{multirow}
\usepackage{hyperref}
\usepackage{rotating}
\usepackage{adjustbox}
\usepackage{threeparttable}
\usepackage{booktabs}

\usepackage{color}
\definecolor{redcolor}{rgb}{1.0,0.,0.}

\definecolor{bluecolor}{rgb}{0,0.,1}

\begin{document}

\preprint{}

\title{Graph machine learning for flight delay prediction due to
    holding manouver}

\author{Jorge L. Franco$^{1,2}$, Manoel V. Machado Neto$^2$, Filipe A. N. Verri$^2$ and Diego R. Amancio$^1$}

\affiliation{
$^1$Institute of Mathematics and Computer Science, University of S\~ao Paulo, S\~ao Carlos, Brazil\\\\
$^2$Aeronautical Technology Institute  -- ITA, S\~ao José dos Campos, SP, Brazil \\
}

\newpage


\begin{abstract}
Flight delays due to holding maneuvers are a critical and costly phenomenon in aviation, driven by the need to manage air traffic congestion and ensure safety. Holding maneuvers occur when aircraft are instructed to circle in designated airspace, often due to factors such as airport congestion, adverse weather, or air traffic control restrictions. This study models the prediction of flight delays due to holding
maneuvers as a graph problem, leveraging advanced Graph Machine
Learning (Graph ML) techniques to capture complex interdependencies in
air traffic networks. Holding maneuvers, while crucial for safety,
cause increased fuel usage, emissions, and passenger dissatisfaction,
making accurate prediction essential for operational
efficiency. Traditional machine learning models, typically using
tabular data, often overlook spatial-temporal relations within air
traffic data. To address this, we model the problem of predicting
holding as edge feature prediction in a directed (multi)graph where we
apply both CatBoost, enriched with graph features capturing network
centrality and connectivity, and Graph Attention Networks (GATs),
which excel in relational data contexts. Our results indicate that
CatBoost outperforms GAT in this imbalanced dataset, effectively
predicting holding events and offering interpretability through
graph-based feature importance. Additionally, we discuss the model's potential operational impact through a web-based tool that allows users to simulate real-time delay predictions. This research
underscores the viability of graph-based approaches for predictive
analysis in aviation, with implications for enhancing fuel efficiency,
reducing delays, and improving passenger experience.
\end{abstract}

\maketitle








\frenchspacing



\doublespacing

\section{Introduction}
\label{Introdução}

\noindent The aviation industry increasingly relies on data-driven
approaches to improve operational efficiency and reduce delays. Among
the pressing challenges in air traffic management is the prediction of
`holding' maneuvers, where aircraft are instructed to delay landing
due to factors such as airport congestion, adverse weather, or
airspace limitations. While holding patterns are necessary for safety,
they contribute to increased fuel consumption, emissions, and
passenger dissatisfaction. This study aims to enhance the accuracy
of holding predictions using machine learning (ML) models based on
graph-structured data, specifically employing advanced methodologies
in Graph Machine Learning (Graph ML) and Graph Neural Networks (GNNs).

Traditional machine learning applications in aviation have primarily
focused on flight delay prediction and air traffic flow
management. For instance, delay predictions based on weather
conditions, airport congestion, and flight schedules have been widely
studied \cite{lambelho2020assessing, gui2019flight}. However, these
models often rely on tabular data representations, which limit their
ability to capture complex relational patterns among airports and
other influencing factors. Additionally, research specifically focused
on holding maneuvers is limited and generally lacks machine learning
and network-based approaches that can model the spatial and temporal
dependencies intrinsic to air traffic data \cite{lee2020development,
  smith2008management}.

The use of graph-based machine learning methods is rapidly advancing
in the field of intelligent transportation, where graph structures
effectively capture complex spatial and temporal relationships across
networks. A recent survey on GNNs in intelligent transportation
systems, \cite{gnnSurveyTransportation}, highlights their
application across a variety of domains, including traffic
forecasting, demand prediction, and urban planning. This survey
underscores the power of GNNs in applications where data is inherently
interconnected, such as autonomous vehicle routing and intersection
management. By organizing studies within these domains, they identify
distinct opportunities and challenges, particularly in multi-modal
models and reinforcement learning applications. Similarly,
\cite{zhao2020tgcn} demonstrates the value of GNNs in the
specific context of real-time traffic forecasting with their T-GCN
(Temporal Graph Convolutional Network) model. By combining Graph
Convolutional Networks (GCN) and Gated Recurrent Units (GRU), the
T-GCN model captures both spatial and temporal dependencies, achieving
state-of-the-art accuracy in urban traffic prediction tasks. These
studies exemplify the increasing role of GNNs in
transportation-related decision-making, showing potential for improved
accuracy and efficiency in complex, dynamic systems.

The study employs two main approaches:

\begin{enumerate}
\item \emph{Tabular-based approach:} We use the CatBoost model,
  leveraging graph features—such as centrality and connectivity
  metrics—that capture the significance of directed edges (flights)
  within the network \cite{prokhorenkova2018catboost}.
\item \emph{Graph Attention Network (GAT) approach:} We compare
  with GATs that have proven effective in applications where
  relational data is essential, making them a promising choice for
  capturing the interconnected nature of air traffic
  \cite{velickovic2017graph}.
\end{enumerate}

The contributions of this study are twofold. First, we demonstrate the application of graph-based ML models for predicting holding events, offering a detailed perspective on airport interdependencies. Second, by comparing the CatBoost model with the GAT, we assess which method better captures the graph topology of air traffic and achieves superior predictive performance. This research has potential implications for improved fuel efficiency, reduced delays, and enhanced passenger experiences by refining model selection and feature engineering strategies tailored to aviation applications. To the best of our knowledge, this is the first application of graph-based ML approaches to address the problem of flight delays caused by holding maneuvers.

The structure of this work is as follows. In Section
\ref{TheoreticalFramework}, we review relevant literature and build
the theoretical background needed. Section \ref{Materials} details the
dataset and the modelling. Section \ref{Results} discusses the
experimental setup, with results and comparison analysis presented and
model deployment. 




\section{Theoretical Framework and Related Works}
\label{TheoreticalFramework}

Graph machine learning can be tracked backwards to the problem of
`learning' on data that is inherently a graph \cite{JMLR:Perozzi} or can be modeled as a graph
\cite{verri2013,grape2020}. This field encompasses a variety of tasks,
including node/edge classification, network construction, link
prediction, graph classification, graph cut/partitioning, network
embeddings, graph coarsening/reduction, which rely on learning
representations from graph-structured data~\cite{correa2019word,amancio2016network,semeraro2025emoatlas,talaga2023polarization,gui2024link,correa2017patterns,morris2024future}. Over the last decades,
researchers have developed numerous approaches to tackle these
challenges, initially these techniques were most developed by complex
networks researchers. However, in the last decade with the
advancements in deep learning, the field has seen a significant shift
towards the merging of three main communities: graph signal
processing, deep learning and complex nets.

As described, defining the field of graph machine learning is not
straightforward, as it encompasses a broad range of methods and
applications. The tasks mentioned above are just a few examples of the
many challenges that can be addressed through graph-based learning
techniques. For clarity, these tasks can be categorized into three
main learning paradigms: supervised, unsupervised, and semi-supervised
learning. In this study, we are interested on the (semi-)supervised
learning paradigm, which encompasses a variety of techniques designed
to leverage learning to (partially) labeled data
\cite{verri2018advantages,amanciof}. But we can refine even more, in
fact, this work will focus in the subset of graph elements
prediction(classification/regression) methods.

In this section, we provide an overview of the theoretical framework
of graph machine learning for node/edge prediction. Here we consider
the division of the field into \texttt{traditional} graph learning and
\texttt{deep} graph learning, where here `traditional' refers to the
machine learning techniques applied to graphs before the advent of
graph neural networks, where standard ML algorithms were applied to
graph data and the topological information measures were encoded as
features together with the tabular data
\cite{costa2007characterization}. This bipartition
is what will pave the way of our explanation, since the last decade
has seen a complex interplay between these two approaches. The field's
evolution can be traced back to when \cite{bruna2013spectral}
introduced one of the first GNN architectures leaned on the theory of
graph signal processing. Concurrently, researchers were developing
node embedding techniques like DeepWalk \cite{perozzi2014deepwalk} and
node2vec \cite{grover2016node2vec}, which bridged traditional and deep
approaches while remaining using complex networks concepts. The
subsequent years saw a surge in GNN architectures, including Graph
Convolutional Networks(GCNs) \cite{kipf2016semi} and GraphSAGE
\cite{hamilton2017inductive}, marking a shift towards more
sophisticated deep learning approaches for graphs and the unification
of the field.

In the following sections, we explain each subset, their theory and
applications, and how they have evolved over time. We also discuss the
challenges and limitations of these methods.

\subsection{Traditional Graph Learning}

\label{classical_learning}

These early efforts focused on shallow learning techniques, such as
feature engineering, graph traversal algorithms, and spectral methods,
which laid the foundation for understanding graph structure and
dynamics. Methods like community detection, centrality measures, and
link prediction became key tools for analyzing
large-scale networks in areas such as social science, biology, and
infrastructure systems
\cite{newman2018networks,boccaletti2006complex}. A significant focus
of these techniques was to develop graph-based features that could be
integrated into traditional machine learning models, effectively
transforming graph data into a format compatible with standard
algorithms like logistic regression, decision trees, and support
vector machines. By encoding graph topology through hand-crafted
features, such as connectivity and centrality, researchers could
leverage these features for tasks like classification, regression, and
clustering in tabular machine learning frameworks.

Among these features, centrality measures became particularly
important due to their ability to capture the relative importance or
influence of nodes in a graph, not just nodes
\cite{bonacich1987power}, but other graph elements such as edges
\cite{Lu2013edgebetw, brohl2019centrality} and hyperedges
\cite{tudisco2021hyperedge}. Centrality measures, such as degree,
betweenness, and closeness, served as input features in machine
learning pipelines, helping to predict outcomes based on the
structural role of nodes within the network.

Spectral centrality, particularly eigenvector centrality
\cite{bonacich1987power}, has proven valuable in machine learning
applications due to its ability to identify globally influential
nodes. Eigenvector centrality assigns a score to each node by
considering not only its direct connections but also the centrality of
its neighbors, which results in a recursive
definition. Mathematically, the eigenvector centrality $x$ of a node
in a graph can be defined as the solution to the equation $Ax = \lambda x$,
where $A$ is the adjacency matrix of the graph, and $\lambda$ is the largest
eigenvalue, thus $x$ is the eigenvector associated with the largest
eigenvalue. This relationship arises from the fact that the centrality
of a node is proportional to the sum of the centralities of its
neighbors, if we normalize the adjacency we get an stochastic matrix
and then $\lambda =1 $ is the largest eigenvalue, named the \texttt{Perron
  vector}. The eigenvector centrality captures both local and global
structure in a network, making it a powerful feature for tasks such as
node classification, ranking, and recommendation systems. A related
and widely used spectral measure is PageRank \cite{brin1998pagerank},
which extends the idea of eigenvector centrality by introducing a
damping factor to model random surfing behavior,
\[ PR(v) = \frac{1 - d}{N} + d \sum_{u \in \mathcal{N}(v)}
  \frac{PR(u)}{\text{deg}(u)},
\] where $PR(v)$ is the PageRank score of node $v$, $d$ is the damping
factor, and $\mathcal{N}(v)$ represents the neighbors of node $v$. This
iterative computation converges to a stationary distribution of
scores, which can be interpreted as the probability of landing on a
given node after a long random walk, in this sense the \texttt{Perron
  vector} signifies the convergence of the process in the
infinite. PageRank has been widely used in ranking tasks, such as
identifying important websites in search engines or recommending
influential users in social networks.

However, these spectral-based centralities come with
limitations. Eigenvector centrality requires the computation of the
principal eigenvector of the adjacency matrix, which involves finding
the largest eigenpair problem. This has a time complexity of $\mathcal{O}(n^2
d)$ for exact methods, where $n$ is the number of nodes in the graph
and $d$ is the ratio of convergence for the power method. Furthermore,
spectral methods can suffer from limitations rooted in the
Perron-Frobenius theorem, which guarantees the existence of a unique
largest eigenvalue only for irreducible, non-negative matrices. For
graphs that are disconnected or have negative weights, these
conditions are violated, and the eigenvector centrality may not be
well-defined or interpretable. That is, the adjacency matrix should be
non-negative and irreducible, where we could use the Perron test $\sum
A^k > 0$ to see if the graph is strongly connected.  These
centralities also tend to be node-centric, lacking a direct extension
to edge importance. For edge centrality, betweenness remains crucial,
particularly in directed graphs, where the structural role of links
(edges) must be considered to capture flow dynamics. Additionally,
spectral centralities can be sensitive to noise and small
perturbations in the graph structure, leading to instability in the
centrality scores. Despite these challenges, spectral centrality
remains a powerful tool for machine learning tasks that benefit from
capturing global graph structure, provided that the computational and
stability issues can be managed.

\subsection{Deep graph learning}

\label{deep_learning}

The rise of deep learning has revolutionized the field of graph
machine learning, enabling the development of more powerful and
scalable models for graph data. Graph neural networks can be divide in
two main categories: spectral-based and spatial-based. Here is a trick
thing, the GCN architecture \cite{kipf2016semi} is commonly divulgated
as a spatial-based method, since it is more intuitive talking about
the convolution operation in the spatial domain, where we simply
aggregate information from the immediate neighbors. However, the GCN
is a spectral-based method, in fact, it can be thought as a
simplification of the first spectral GNN \cite{bruna2013spectral}
proposed and that builds the math behind GCNs. That said, first we
introduce the spectral-based GNNs and then the spatial-based ones.

\subsubsection{Spectral-based GNNs}

\label{spectral-based}

Spectral methods are rooted in graph signal processing. The core idea
is that a signal on a graph can be represented as node features, where
each feature vector at a node corresponds to a `signal' defined over
the graph. In this context, the graph Laplacian $\mathcal{L} = D - A$, where $D$
is the degree matrix and $A$ is the adjacency matrix, plays a crucial
role. It captures the structure of the graph and can be used to
perform operations analogous to Fourier transforms in classical signal
processing. Spectral methods can be categorized into two types:
eigenvalue-based, where the focus is on creating a graph filter in the
Fourier domain, and eigenvector-based, where the goal is to use a
spectral basis to decompose the signal
\cite{bo2023surveyspectralgraphneural}.

\cite{bruna2013spectral} introduced the first spectral Graph
Neural Network (GNN), termed the Spectral CNN (SCNN), which aimed to
translate ideas from standard Convolutional Neural Networks for images
to graphs. The SCNN leverages the spectral decomposition of the graph
Laplacian $\mathcal{L} = U \Lambda U^T$ to define a filter convolution operation in
the Fourier domain. In this framework, the graph Fourier transform of
a signal $f$ is represented as $\hat{f} = U^T f$, and the convolution
operation ($\star$) is defined as $g_{\theta} \star f = U g_{\theta} U^T f$, where
$g_{\theta}$ is a learnable filter parameterized by $\theta$. While powerful,
the SCNN faces significant challenges: it requires $\mathcal{O}(n^3)$
computational complexity to calculate the entire graph spectrum, which
is prohibitively expensive for large graphs. Moreover, the
non-localized nature of eigenvectors means global information can
overshadow local structural details, leading suboptimal balance
between local and global information aligned with a huge parameter
complexity \cite{usgnn}.

To address these limitations,
ChebNet~\cite{defferrard2016convolutional} introduced Chebyshev
polynomials to approximate spectral filters, effectively reducing
computational complexity while preserving the ability to capture
localized patterns in the graph structure. The main ideia is to
redefine our previous filtering operation to $ g_{\theta}(\mathcal{L} ) f =
\sum_{k=0}^{K-1} \theta_k T_k(\widetilde{\mathcal{L}}) f $, where $T_k(\widetilde{\mathcal{L}}) =
$ is the Chebyshev polinomial of order k evaluated at the scaled
Laplacian $\widetilde{\mathcal{L}} = 2 \frac{\mathcal{L}}{\lambda_\text{max}} - I_n$. This
innovation not only makes spectral GNNs more scalable to larger
graphs, since we just need to calculate the first eigenpair ($\mathcal{O}(n^2)$
through the power method) for the approximations, but also enhances
their ability to balance local and global information processing. In
fact, the filters are $K$-localized for polinomials of order $K$, that
is intuitive by remembering that $\mathcal{L} ^K$ represents the paths with
length less or equal to $K$.  The ChebNet laid the foundation for GCNs
\cite{kipf2016semi}. Although GCNs are commonly referred to as spatial
methods, their underlying principle is rooted in the truncation of the
Chebyshev expansion to $K=1$, which limits the filter to first-order
neighbors. This simplification reduces computational complexity
significantly while preserving effectiveness. Instead of requiring the
full spectral decomposition of the Laplacian matrix, GCNs use a
localized approximation of the graph convolution, expressed as: $g_{\theta}
\star f \approx \theta (I_n + \widetilde{A}) f$, where $\widetilde{A} =
D^{-\frac{1}{2}} A D^{-\frac{1}{2}}$ is the normalized adjacency
matrix, where $A$ is the adjacency matrix, and $D$ is the degree
matrix. This approximation results in an efficient propagation rule
that aggregates information from a node's immediate neighbors while
updating the node's features. This propagation mechanism is often
confused as a spatial method because it effectively propagates
information from adjacent nodes—akin to a spatial neighborhood
aggregation. Although its already a simple model, results have shown
that GCNs can achieve state-of-the-art performance on a variety of
tasks with even more simplifications \cite{wu2019simplifying}.
However, as we can note, all these spectral methods works just in
undirected graphs, since it needs the spectral
decomposition. Furthermore, these methods are `node centric', that is,
they focus just on node features and the topology of the nodes, most
of this is because the adjacency matrix maps the dimension of nodes to
nodes, thus leaving \emph{edge features} out of the scene.

\subsubsection{Spatial-based GNNs}

\label{spatial-based}

Spatial-based GNNs differ from spectral-based approaches by directly
leveraging the graph structure to perform convolutions in the spatial
domain, rather than relying on the spectral decomposition of graph
operators like the Laplacian. In spatial-based methods, the
convolution operation is interpreted as an aggregation of node
features from a node's local neighborhood, akin to how standard
convolutional neural networks aggregate pixel information from nearby
regions in image data. These methods operate by iteratively updating
node representations by propagating information between neighboring
nodes, making them intuitive and highly scalable for large-scale
graphs.

The general framework for message passing in spatial-based GNNs can be
described as follows. For each node $i$, at layer $t$, we aggregate
the features of its neighbors $\mathcal{N}(i)$ to produce an updated node
embedding: $\mathbf{m}_i^{(t+1)} = \text{AGGREGATE}^{(t)} \left(
  \left\{ \mathbf{h}_j^{(t)} : j \in \mathcal{N}(i) \right\} \right)$, where
$\mathbf{h}_j^{(t)}$ is the feature of neighboring node $j$ at layer
$t$. Then, we update the node $i$'s representation:
$\mathbf{h}_i^{(t+1)} = \text{UPDATE}^{(t)} \left( \mathbf{h}_i^{(t)},
  \mathbf{m}_i^{(t+1)} \right)$, where $\text{AGGREGATE}^{(t)}$ is a
neighborhood aggregation function, and $\text{UPDATE}^{(t)}$ is the
node update function.

The general idea behind spatial-based GNNs is that, for each node, we
aggregate the features of its neighbors to produce an updated node
embedding. A key example of this is the GraphSAGE architecture
\cite{hamilton2017inductive}, which computes node representations by
sampling and aggregating features from the node's neighbors. The
GraphSAGE model employs several types of aggregation functions,
including mean, LSTM-based, and pooling aggregators, which allow for
flexible and inductive learning on large graphs. In particular,
GraphSAGE enables the generation of embeddings for unseen nodes,
making it suitable for inductive learning tasks, where the model needs
to generalize to new nodes that were not present during
training. Unlike spectral-based methods, which are constrained to a
fixed graph size and structure due to their reliance on the graph
Laplacian, spatial-based GNNs are inherently more flexible and can be
applied to dynamic and evolving graphs. These models perform
neighborhood aggregation locally, and therefore do not require the
global knowledge of the graph structure that spectral methods
need. This flexibility makes them particularly useful for large-scale
graphs and for graphs where the structure may change over time, such
as social networks or knowledge graphs.

Another prominent spatial-based GNN is the Graph Attention Network
(GAT) \cite{velickovic2017graph}, which introduced attention
mechanisms into graph learning. GAT models learn to assign different
weights to the neighbors of a node, allowing the model to focus more
on the most relevant neighbors during the feature aggregation
process. This is achieved using a self-attention mechanism, where the
importance of neighboring nodes is learned through a shared attention
coefficient, $ e_{ij} = \text{LeakyReLU}(\mathbf{a}^T [\mathbf{W}
\mathbf{h}_i || \mathbf{W} \mathbf{h}_j]) $, where $e_{ij}$ represents
the attention coefficient between nodes $i$ and $j$, $\mathbf{W}$ is a
learnable weight matrix, $\mathbf{h}_i$ and $\mathbf{h}_j$ are the
feature vectors of nodes $i$ and $j$, and $||$ denotes
concatenation. The attention coefficients are then normalized across
all of a node's neighbors using the softmax function, $ \alpha_{ij} =
\frac{\exp(e_{ij})}{\sum_{k \in \mathcal{N}(i)} \exp(e_{ik})} $, this normalization
ensures that the attention coefficients sum to 1, allowing the model
to perform a weighted aggregation of the neighbors' features, $
\mathbf{h}_i' = \sigma \left( \sum_{j \in \mathcal{N}(i)} \alpha_{ij} \mathbf{W} \mathbf{h}_j
\right)$, here $\mathbf{h}_i'$ is the updated representation of node
$i$, and $\sigma$ is a non-linear activation function. By learning
attention coefficients, GATs can capture both the importance and the
structure of the graph, making them particularly effective in tasks
where the relationships between nodes are not equally important, such
as in citation networks or social media graphs.




\section{Materials and Methods}
\label{Materials}

This section details the materials and methods used in our study, providing a fluid overview of the datasets and aviation features employed in our predictive modeling. Our objective is to predict whether a given aircraft will experience a delay due to a holding maneuver by leveraging both a CatBoost model enhanced with graph-derived features and a Graph Attention Network (GAT) approach.

The study utilizes two distinct datasets, each comprising 42,336 observations that encompass a comprehensive range of meteorological, geographical, and flight-specific features. These datasets were constructed from weather reports and airport/flight specifications provided by ICEA, and they are tailored for different predictive tasks. In the binary classification dataset, the model is tasked with predicting the occurrence of a holding maneuver. Here, the vast majority of samples (41,616) represent flights without holding delays, while a small subset (720) corresponds to flights with holding delays. This significant imbalance challenges the model to accurately predict a rare event without succumbing to overfitting. In contrast, the regression dataset is designed to predict the precise duration of holding delays, with each observation labeled by a continuous value representing the delay in seconds. This approach offers a more detailed analysis of delay durations, enhancing the potential for operational insights.

Aviation features in our datasets can be understood through three intertwined dimensions. First, meteorological features—sourced from both METAR and METAF reports—provide real-time weather observations and complementary forecast elements. METAR (Meteorological Terminal Aviation Routine Weather Report) delivers data on wind direction and speed, visibility, temperature, and cloud coverage, while METAF reports offer additional context that deepens the understanding of atmospheric conditions that might influence holding maneuvers. Second, geographical features offer spatial context by capturing the geodesic flight distance between departure and arrival airports, along with details such as airport altitudes and precise latitude and longitude coordinates. This spatial information is critical for analyzing how physical location and distance affect delay events. Finally, flight-specific features capture operational nuances such as the flight hour and indicators of runway activity, including previous runway head changes and recent alterations in runway configuration. Together, these features provide a holistic view of the operational environment, allowing for a more nuanced prediction of holding maneuvers.

In our modeling approaches, we first apply CatBoost—a gradient boosting decision tree model enhanced with graph-derived features from the airport network—which is well-suited to handle class imbalances and offers transparency through explainable AI techniques. In parallel, we explore Graph Attention Networks (GATs), which use attention mechanisms to model the intricate relationships within graph-structured data. Although GATs are promising in capturing relational patterns, our experiments reveal that they struggle with class imbalance, often leading to overfitting in deeper architectures.

The following sections provide further details on the specific implementations and experimental configurations for both the CatBoost and GAT models.

\subsection{CatBoost with Graph Features}
\label{sec:catboost_model}

This study employs the CatBoost model, a high-performance gradient
boosting library, chosen specifically for its ability to handle
categorical features and class imbalance effectively, as well as for
its robust handling of noisy data
\cite{prokhorenkova2018catboost}. CatBoost has been widely recognized
for its superior performance in structured data problems, particularly
when compared to other boosting algorithms like XGBoost and LightGBM,
thanks to its unique techniques such as ordered boosting and
categorical feature encoding. These innovations help prevent
overfitting and enhance generalization in class unbalanced problems.

Here, we describe how CatBoost is combined the graph-based features
that are extracted of our modeled airports network. These features are
derived from the weighted directed graph and enconded as tabular
features that are used as input to the model as we describe in the
following sections.

\subsubsection{Graph Representation of the Flight Network}

\begin{figure}[h] 
    \centering
    \includegraphics[width=0.3\textwidth]{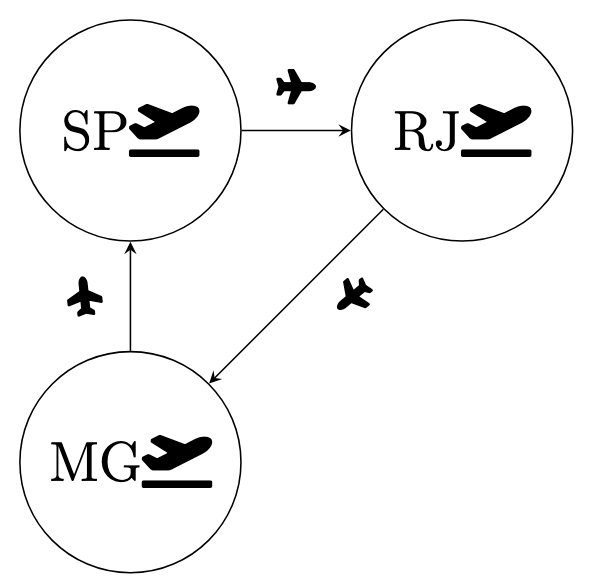} 
    \caption{Example of airports and directed flights.}
    \label{fig:flight_network_graph}
\end{figure}
\begin{figure}[h]
  \centering
  \includegraphics[width=0.8\textwidth]{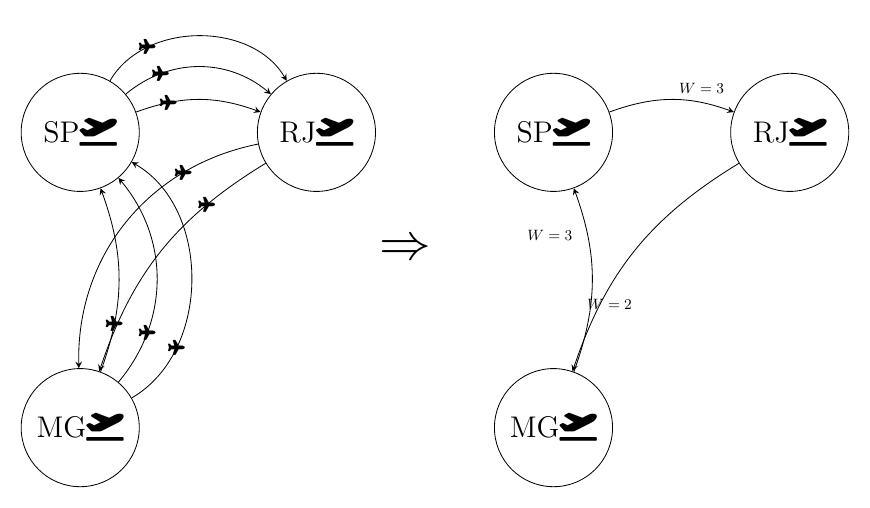}
  \caption{Transformation of a multigraph of flights into a weighted directed graph. The multigraph (left) represents multiple flights between airports. In the weighted graph (right), edges are aggregated to show total flights as weights.}
  \label{fig:multigraph_to_weighted_graph}
\end{figure}

To model the interactions in flight data, we represented the problem
as a directed graph, depicted in
Figure~\ref{fig:flight_network_graph}, where each node represents an
airport, here we represent the airports as states of Brazil: SP (São
Paulo), MG (Minas Gerais), RJ (Rio de Janeiro). In this network, nodes represent airports. Directed edges represent flights, with each edge directed from the departure airport to the destination airport.
Given the frequent occurrence of multiple flights
between the same pairs of airports (i.e., multiedges), we have in fact
a multigraph, however we abstract it into a weighted directed graph as
shown in \ref{fig:multigraph_to_weighted_graph}. Here, each edge's
weight corresponds to the total number of flights between a specific
pair of airports, transforming multiple directed edges into a single
weighted edge. This abstraction allows us to calculate key network
metrics more easily, which we then used as features in the CatBoost
model.

\subsubsection{Graph-based Features} The graph-based features encode
essential structural information about the flight network, capturing
connectivity, centrality, and robustness. These features are crucial
for understanding the influence of each airport within the network and
its potential impact on flight holding patterns.

Although we have already made this simplification of the multigraph,
transforming it into a weighted directed graph, we still need to
extract the features from the graph and encode them as tabular
data. However, this is not straightforward, as the graph measures are
not directly compatible with the model.

The modelling will impact dramatically in the resulting graph-based
features. For instance, we need to calculate edge measures, but this
is not so explored as node measures, so the lack of possibilities is a
challenge to be overcome.  Another challenge is the direction, that
is, we have to create edge measures in a directed weighted graph,
which is hard, as we detailed in section \ref{classical_learning},
because most of the complex networks measures proposed are `node
centric' and for undirected graphs.

With this in mind, we can observe why the weighted graph
transformation was so important, since the measures available for our
setting are strongly dependent to the weight (as we will detail
later), and our graph is almost totally connected, so in undirected
unweighted setting they would be approximately equal, leaving no
information. The following graph metrics were calculated from the
weighted directed graph:

\begin{enumerate}
\item \emph{Betweenness Centrality:} Captures the relative
importance of each airport in terms of the routes it controls within
the network. Higher values indicate airports that serve as critical
transit points.
\item \emph{Flow Betweenness:} Highlights the flow dynamics of
connections, showing how flights tend to route through certain
airports, which may correlate with congestion.
\item \emph{Edge Connectivity:} Indicates the robustness of airport
connections, with higher values signifying more resilient routes
between airports that could better handle rerouting needs.
\item \emph{Degree Difference:} Measures the disparity between
in-degrees and out-degrees at each node, helping to identify key hubs
or spokes in the network.
\item \emph{Google Matrix:} Based on PageRank centrality, the Google
matrix provides a probabilistic transition representation for each
airport, which reflects both local and global connectivity.
\end{enumerate}

As we can see, these features are not commonly used in the
literature. Here is where the weighted network plays a crucial role,
edge betweeness centrality \cite{newman2004finding} is constructed
using shortest paths in the network, thus the weight will be crucial
part of it, since without it the graph is almost fully connected, the
shortest path will be almost the same for all pairs of nodes. The same
happens with flow betweeness centrality \cite{freeman1991centrality},
that is a measure based on electrial circuits Kirchoff law, more
specifically, instead of working with shortest paths, it use the
maximum flow that pass through each edge and the weight visualized as
capacity will be crucial to calculate it.

The edge connectivity is a measure of the minimum number of edges that
must be removed to disconnect the graph, and the weight will be
crucial to calculate it. The degree difference we stated here as a
measure of the difference between the in-degree and out-degree of a
node. The Google matrix is a way we derived to keep using PageRank for
edges. In fact, as we detailed in section \ref{classical_learning},
althought the PageRank centrality could be applied in our graph, since
it satisfieis the Perron theorem as it is always postivie and strongly
connected, it is a node measure, so we have to adapt it to edges, and
the Google matrix is a way to do it.

These features enhance the CatBoost model by embedding graph-theoretic
insights into its predictive capabilities, ultimately enabling a more
nuanced understanding of how network dynamics relate to flight holding
patterns.

\section[Graph Attention Network]{Graph Attention Network}
\label{Graph Attention Network}

As we previously described, the GAT model in section
\ref{spatial-based} has a large range of applications, from drug
discovery to fake news detection \cite{keywordsCaravanti}. The GAT
model leverages the underlying graph structure but does not rely on
explicitly computed graph-derived features like the CatBoost model
does. Instead, it learns node representations in an end-to-end manner,
enabling the model to capture the relationships between airports and
flights directly from the data.

The modelling of a GNN for our problem is a challenging task, as we
have to adapt the model to predict edge features, since `holding' is
an edge feature in our setting. In section \ref{spectral-based} we
detailed why the spectral-based GNNs are not suitable for our setting,
as they are not able to handle edge features and direction, due to
their `node-centric' approach based on the adjacency matrix. Although
spatial-based GNNs can handle direction in their majority, they are
not able to handle edge features in general, since they need to create
a way to aggregate the edge features with the neighbors' features.

The GAT model is so used because it is highly adaptable in pratically
any graph setting. As we will show, the attention mechanism detailed
in section \ref{spatial-based} can be generalized to handle edge
features, and the model can be adapted to predict edge features. In
fact, a simple concatenation ($ || $) in the attention formula already
gives us this power,
\begin{equation}
 \alpha_{ij} = \sigma(\phi_1( \mathbf{a}^T [ W h_i || W h_j || W_2 e_{ij} ]))  \text{,}
\end{equation}
where $e_{ij}$ are the edge features, $h_i$ and $h_j$ are the node
  features, and $W$ and $W_2$ are the weight matrices. This formula
  allows the model to focus on the relevant neighboring nodes, making
  it ideal for relational data. In our case, the edge features are the
  tabular data features with holding being part of them, which is the
  target we want to predict. 

\begin{figure}
    \centering
    \includegraphics[width=0.9\linewidth]{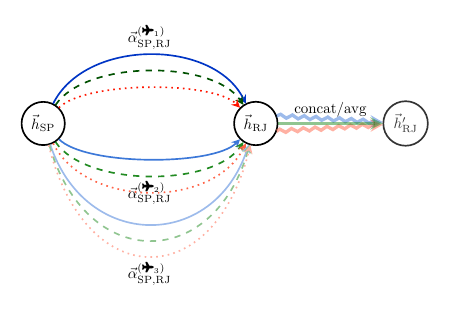}
    \caption{Airport multigraph GAT Layer with multi-head attention for three different flights between nodes (SP,RJ), with alternating colors opacity for each flight.}
    \label{fig:multigraph_layer}
\end{figure}

Furthermore, the directed multigraph setting we described in section
\ref{sec:catboost_model} is not a problem for the GAT model, since it
can handle multiple edges between the same pair of nodes, as we will
show in the following sections. We show how we model the GAT to be a
directed multigraph representing the flights and their features in
Figure \ref{fig:multigraph_layer}.


Finally, the last layer of our predictor would be to pass the learned
node embeddings $h_i$ and $h_j$ with the edge feature $e_{ij}^{(k)}$
of the flight $k$ to a fully connected layer (MLP) to predict the
holding of the flight $k$.  That is, we simply concatenate them, and
after the MLP layer, we have a sigmoid $\sigma$ activation function that
outputs the prediction $\hat{y}_k$ of holding,

$$ \hat{y}_k = \sigma (\text{MLP}(h_i || h_j || e_{ij}^{(k)})) \; \; \; \text{.} $$

\section{Results}
\label{Results}

This section presents a comprehensive evaluation of the models applied in this study, with a focus on comparing the CatBoost and Graph Attention Network (GAT) models, as well as examining the regression capabilities and interpretability of the CatBoost approach. The discussion that follows preserves detailed narrative paragraphs while reducing the number of sections and removing the step-by-step definitions of performance metrics.

\subsection{Model Performance and Interpretability}
\label{sec:model_performance}

The performance of each model in predicting flight delays due to holding maneuvers was evaluated using a range of metrics, including accuracy, precision, recall, and F1-score. Table \ref{tab:gat_catboost_metrics} summarizes the results for various GAT layer configurations alongside the CatBoost model. While the single-layer GAT model achieved the highest accuracy, its precision and F1-score were markedly low. In contrast, the CatBoost model provided a balanced performance across all metrics, which is particularly beneficial for the imbalanced dataset at hand. This balanced performance indicates that CatBoost is more effective at capturing both delayed and non-delayed flights, even if it sometimes errs on the side of caution in predicting delays.

\begin{table}[!htbp]
  \centering
  \caption{Performance metrics for various GAT layer configurations and CatBoost with graph features.}
  \label{tab:gat_catboost_metrics}
  \begin{tabular}{ccccc}
    \toprule
    \textbf{Model} & \textbf{Test Accuracy} & \textbf{Precision} & \textbf{Recall} & \textbf{F1-Score} \\
    \midrule
    \textbf{CatBoost} & 0.90 & \textbf{0.09} & 0.58 & \textbf{0.16} \\
    \textbf{1 GAT Layer} & \textbf{0.95} & 0.03 & 0.06 & 0.04 \\
    \textbf{3 GAT Layers} & 0.52 & 0.01 & 0.40 & 0.03 \\
    \textbf{5 GAT Layers} & 0.57 & 0.01 & 0.30 & 0.02 \\
    \textbf{10 GAT Layers} & 0.91 & 0.02 & 0.08 & 0.03 \\
    \textbf{30 GAT Layers} & 0.02 & 0.02 & \textbf{0.99} & 0.03 \\
    \bottomrule
  \end{tabular}%
\end{table}

\begin{figure}[!htbp]
    \centering
    \includegraphics[width=1\textwidth]{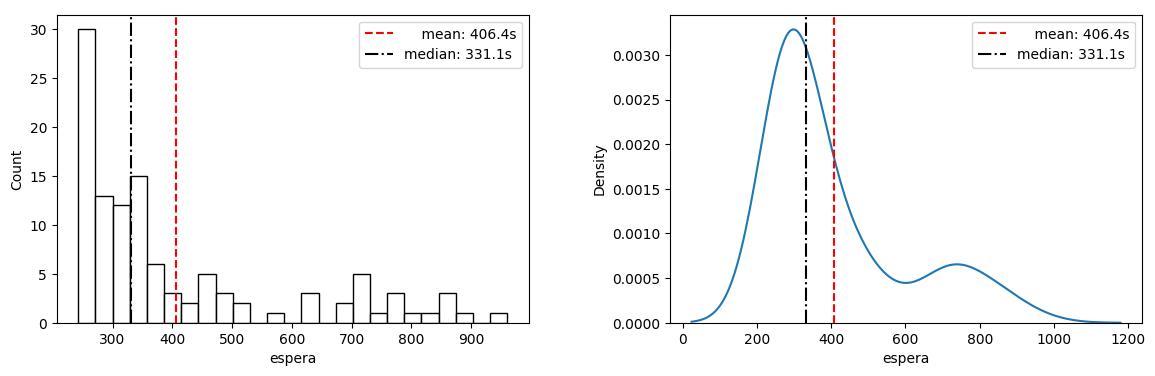}
    \caption{Distribution of predicted delay values ($y_\text{pred}$) for the regression task using CatBoost.}
    \label{fig:y_pred}
\end{figure}

\begin{figure}[!htbp]
    \centering
    \includegraphics[width=1\textwidth]{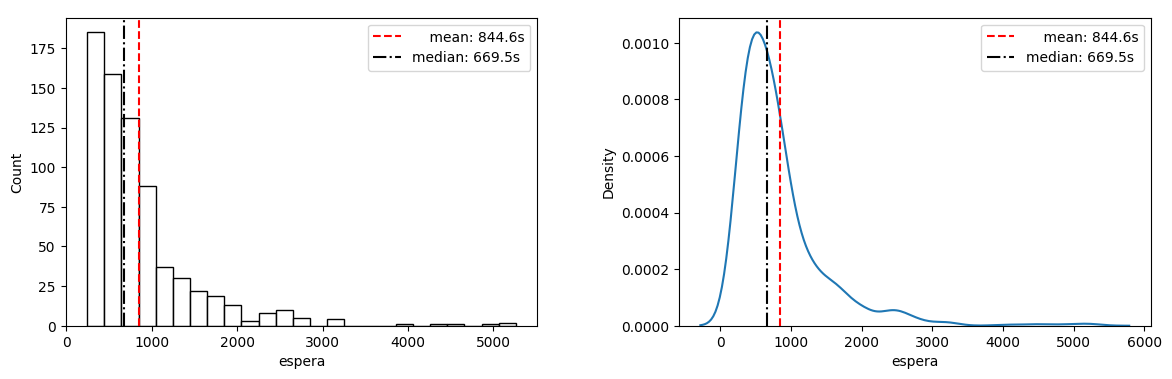}
    \caption{Distribution of actual delay values ($y_\text{test}$) for the test set in the regression task.}
    \label{fig:y_test}
\end{figure}

Beyond classification, the CatBoost model was also applied to a regression task aimed at predicting continuous delay values. Figures \ref{fig:y_pred} and \ref{fig:y_test} illustrate the distributions of the predicted and actual delay values, respectively. The overall alignment between these distributions suggests that CatBoost is effective at capturing the underlying trends in flight delay data, though deviations at extreme values highlight potential areas for further refinement.

\begin{figure}[!htbp]
    \centering
    \includegraphics[width=0.85\textwidth]{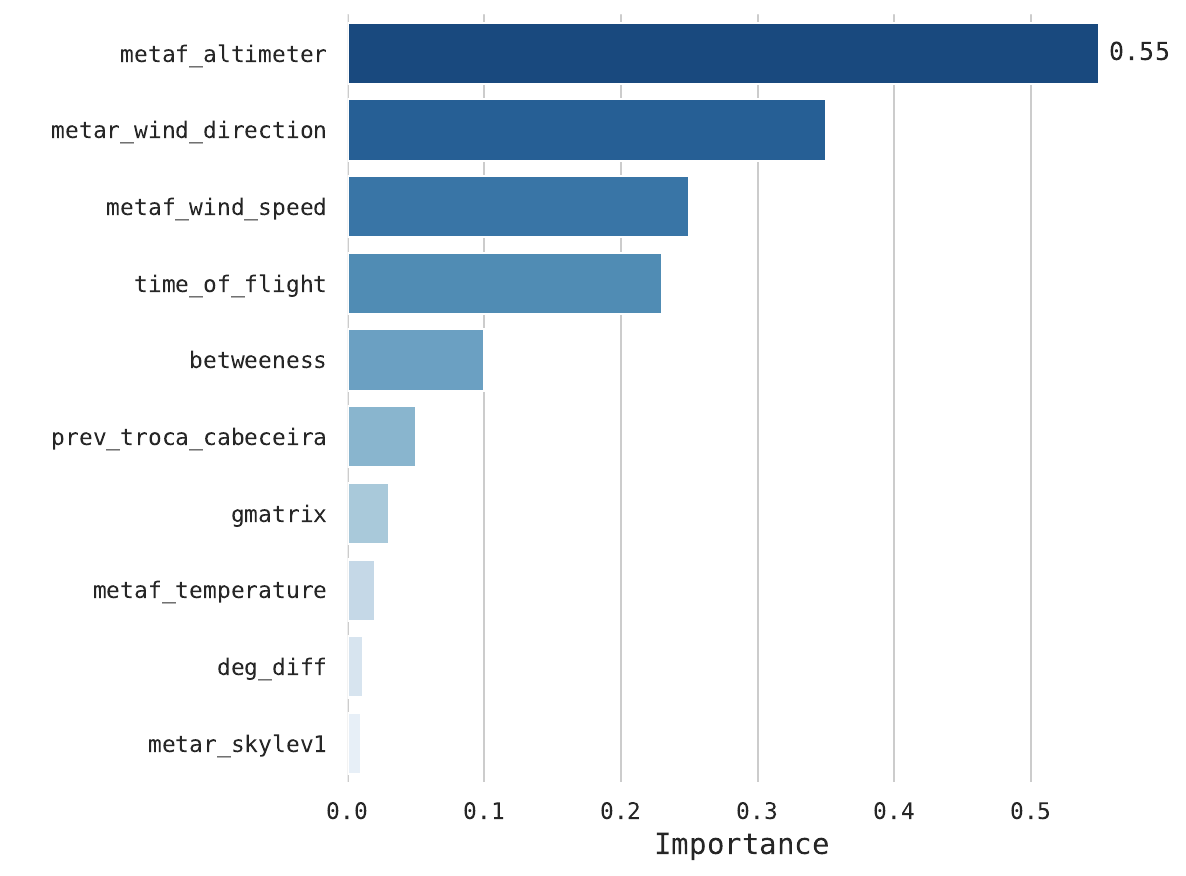}
    \caption{Feature importance for the CatBoost model on the airport network dataset, highlighting the relevance of graph-based features.}
    \label{fig:catboost_feature_importance}
\end{figure}

Interpretability is another key strength of the CatBoost approach. By leveraging graph-based features, the model not only achieves robust performance but also offers transparency regarding feature importance. As shown in Figure \ref{fig:catboost_feature_importance}, Explainable AI (XAI) techniques help elucidate how specific graph-based metrics contribute to the model's predictions, thereby providing valuable insights into the decision-making process.

\subsection{Deployment, Implementation, and Discussion}
\label{sec:deployment_discussion}

The practical applicability of our approach is further demonstrated through the development of a web-based simulation tool. The source code for this project is available on GitHub at \url{https://github.com/graph-learning-ita/airnet-holding-ml/}. The simulation tool, named \textit{Airdelay}, is implemented using Folium and Streamlit and is accessible at \url{https://airdelay.manoel.dev}. This tool visualizes flight delays as predicted by the CatBoost model in real time, enabling users to simulate various operational scenarios and assess the potential impact of holding maneuvers.

\begin{figure}[!htbp]
    \centering
    \includegraphics[width=1\textwidth]{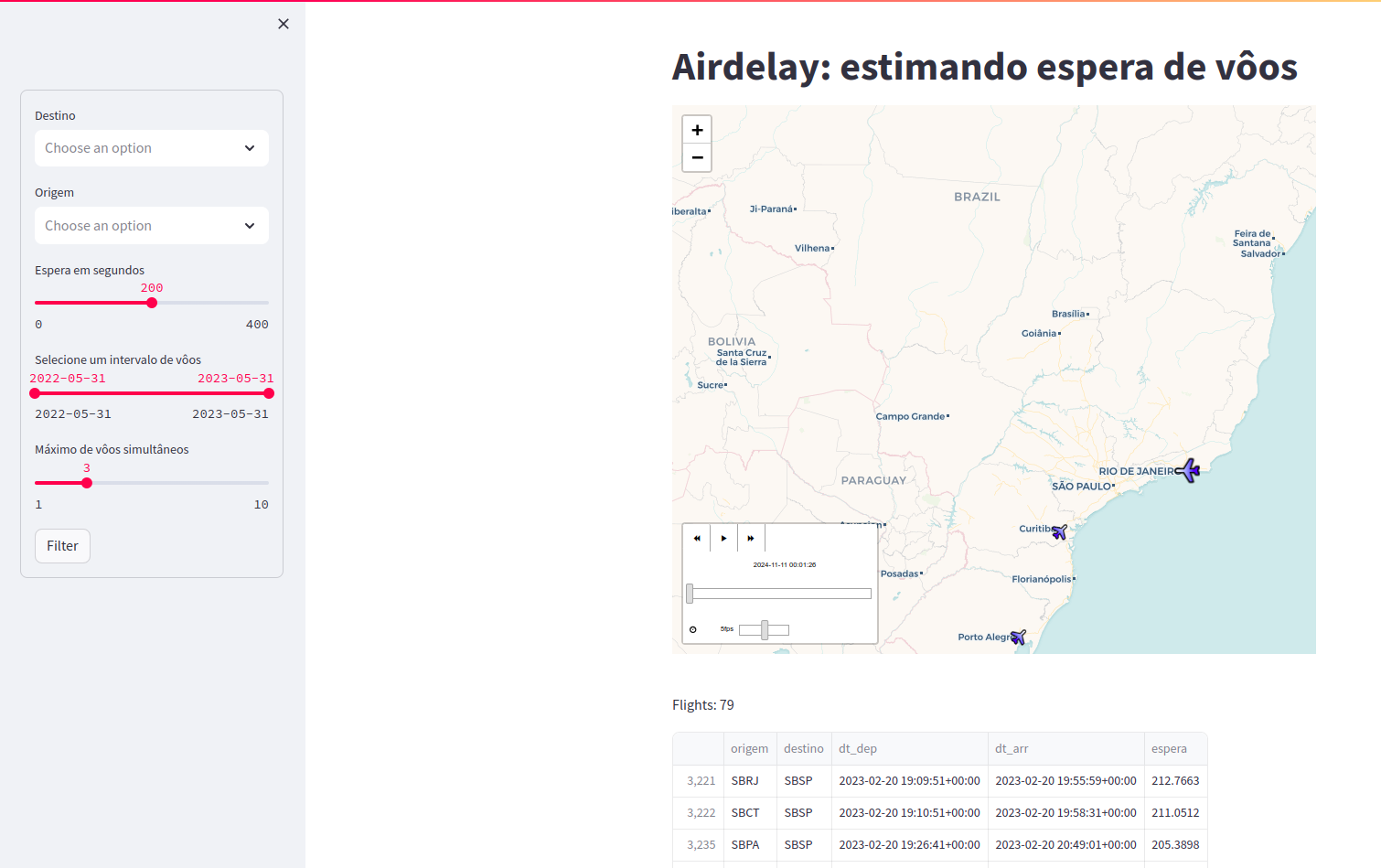}
    \caption{\textit{Airdelay} web-based simulation tool, showing predicted flight delays due to holding maneuvers.}
    \label{fig:airdelay}
\end{figure}

Figure \ref{fig:airdelay} shows a screenshot of the \textit{Airdelay} interface, which allows users to interact with the map, explore different scenarios, and gain a deeper understanding of the model's performance through dynamic visualizations.

In summary, our results underscore the advantages of the CatBoost model for this application. Despite the high accuracy observed with some GAT configurations, issues such as overfitting and unstable performance—particularly in terms of precision and F1-score—limit their effectiveness in an imbalanced setting. The CatBoost model, enhanced by graph-based features, not only achieves superior performance in both classification and regression tasks but also offers interpretability through XAI techniques.

Nevertheless, our study is not without limitations. The scope of hyperparameter tuning was restricted, and the challenges posed by class imbalance remain significant. Future research should focus on refining the GAT architecture, exploring additional graph-based features, and potentially incorporating alternative models such as Support Vector Machines (SVM) to further improve predictive performance. Moreover, emerging GNN architectures, such as the provably powerful graph neural network for directed multigraphs introduced in \cite{egressy2024provably}, hold promise for addressing class imbalance issues and enhancing minority-class performance.

\section{Conclusion }
\label{Conclusion}

This study examined the prediction of flight delays caused by holding maneuvers using graph machine learning techniques. By modeling air traffic as a directed network, we applied both CatBoost, which integrates graph-based features, and Graph Attention Networks (GATs), which capture relational dependencies. Our results demonstrate that while GATs provide a flexible framework for learning from structured data, CatBoost achieved better performance on this imbalanced dataset, highlighting the effectiveness of graph-based feature engineering in structured tabular models. Additionally, our web-based simulation tool illustrates the practical implications of these predictive models for real-time decision-making in air traffic management.

Despite these advances, several areas remain open for improvement. One limitation of our approach is the lack of explicit data imputation techniques for handling missing information. Future work could explore oversampling methods such as GraphSMOTE \citep{zhao2021graphsmote} to enhance the representation of underrepresented holding events in the dataset. Furthermore, since this is an edge-based classification problem, incorporating topological deep learning techniques could offer new insights by leveraging higher-order structures in the flight network. Methods from topological graph learning, such as persistent homology-based representations \citep{immonen2024going}  or topological message passing \citep{bodnar2021weisfeiler} techniques tailored for edge features, may provide a richer understanding of the underlying flight delay dynamics.

Additionally, future research could explore hybrid models that combine GNNs and gradient boosting to leverage both graph-based feature learning and tabular-based decision trees. The integration of real-time air traffic and weather data could further enhance predictive accuracy, making these models more robust for operational deployment. As airspace management continues to evolve, these approaches hold promise for optimizing flight scheduling, reducing delays, and improving overall passenger experience.

\section*{Acknowledgments}

This study was financed in part by the Coordenação de Aperfeiçoamento de Pessoal de Nível Superior - Brasil (CAPES) - Finance Code 001. Diego R. Amancio acknowledges financial support from CNPq (Grant no. 311074/2021-9)  and FAPESP (Grant no. 2020/06271-0).

\newpage

\bibliographystyle{ieeetr}
\bibliographystyle{abbrv}
%

\end{document}